\documentclass[letterpaper]{article} 
\usepackage{aaai2026}  
\usepackage{times}  
\usepackage{helvet}  
\usepackage{courier}  
\usepackage[hyphens]{url}  
\usepackage{graphicx} 
\usepackage{natbib}  
\usepackage{caption} 
\usepackage{algorithm}
\usepackage{algorithmic}
\usepackage{amsfonts}
\usepackage{amsmath}   
\usepackage{amsthm}    
\usepackage{amssymb}   
\usepackage{booktabs}
\usepackage{pifont}
\usepackage[table]{xcolor}  
\usepackage{pifont}
\newcommand{\cmark}{\ding{51}} 
\newcommand{\xmark}{\ding{55}} 
\usepackage{enumitem}
\theoremstyle{plain}

\usepackage{newfloat}
\usepackage{listings}
\DeclareCaptionStyle{ruled}{labelfont=normalfont,labelsep=colon,strut=off} 
\floatstyle{ruled}
\newfloat{listing}{tb}{lst}{}
\floatname{listing}{Listing}
\title{Learning to Deliberate: Meta-policy Collaboration\\for Agentic LLMs with Multi-agent Reinforcement Learning}

\author{
    Wei Yang and Jesse Thomason
}
\affiliations{
    University of Southern California\\

    Los Angeles, CA, USA\\
    \{wyang930, jessetho\}@usc.edu
}

\usepackage{bibentry}

\begin{document}

\maketitle

\begin{abstract}
Multi-agent systems of large language models (LLMs) show promise for complex reasoning, but their effectiveness is often limited by fixed collaboration protocols. 
These frameworks typically focus on macro-level orchestration while overlooking agents’ internal deliberative capabilities. 
This critical meta-cognitive blindspot treats agents as passive executors unable to adapt their strategy based on internal cognitive states like uncertainty or confidence. 
We introduce the Meta-Policy Deliberation Framework (MPDF), where agents learn a decentralized policy over a set of high-level meta-cognitive actions: Persist, Refine, and Concede. 
To overcome the instability of traditional policy gradients in this setting, we develop SoftRankPO, a novel reinforcement learning algorithm. 
SoftRankPO stabilizes training by shaping advantages based on the rank of rewards mapped through smooth normal quantiles, making the learning process robust to reward variance.
Experiments show that MPDF with SoftRankPO achieves a 4-5\% absolute gain in average accuracy across six mathematical and general reasoning benchmarks compared to state-of-the-art heuristic and learning-based multi-agent reasoning algorithms.
Our work presents a paradigm for learning adaptive, meta-cognitive policies for multi-agent LLM systems, shifting the focus from designing fixed protocols to learning dynamic, deliberative strategies.
\end{abstract}



\section{Introduction}

As the limitations of single-agent reasoning become apparent in complex tasks \cite{park2023generative,hong2023metagpt}, the research frontier is  shifting toward multi-agent systems (MAS), where multiple large language model (LLM) agents engage in collaborative reasoning to improve accuracy and robustness \cite{chen2025mao,ping2025verimoa,yang2025maestro}. Their effectiveness often depends on well-designed collaboration protocols, such as multi-turn debates and peer review mechanisms \cite{du2023improving,liu2024groupdebate}. 
However, static protocols prevent
agents from adapting their behavior to the unique demands of a given problem and to evolving context \cite{du2024survey,yue2025masrouter}. 

\begin{figure}[t]
\centering
\includegraphics[width=\columnwidth]{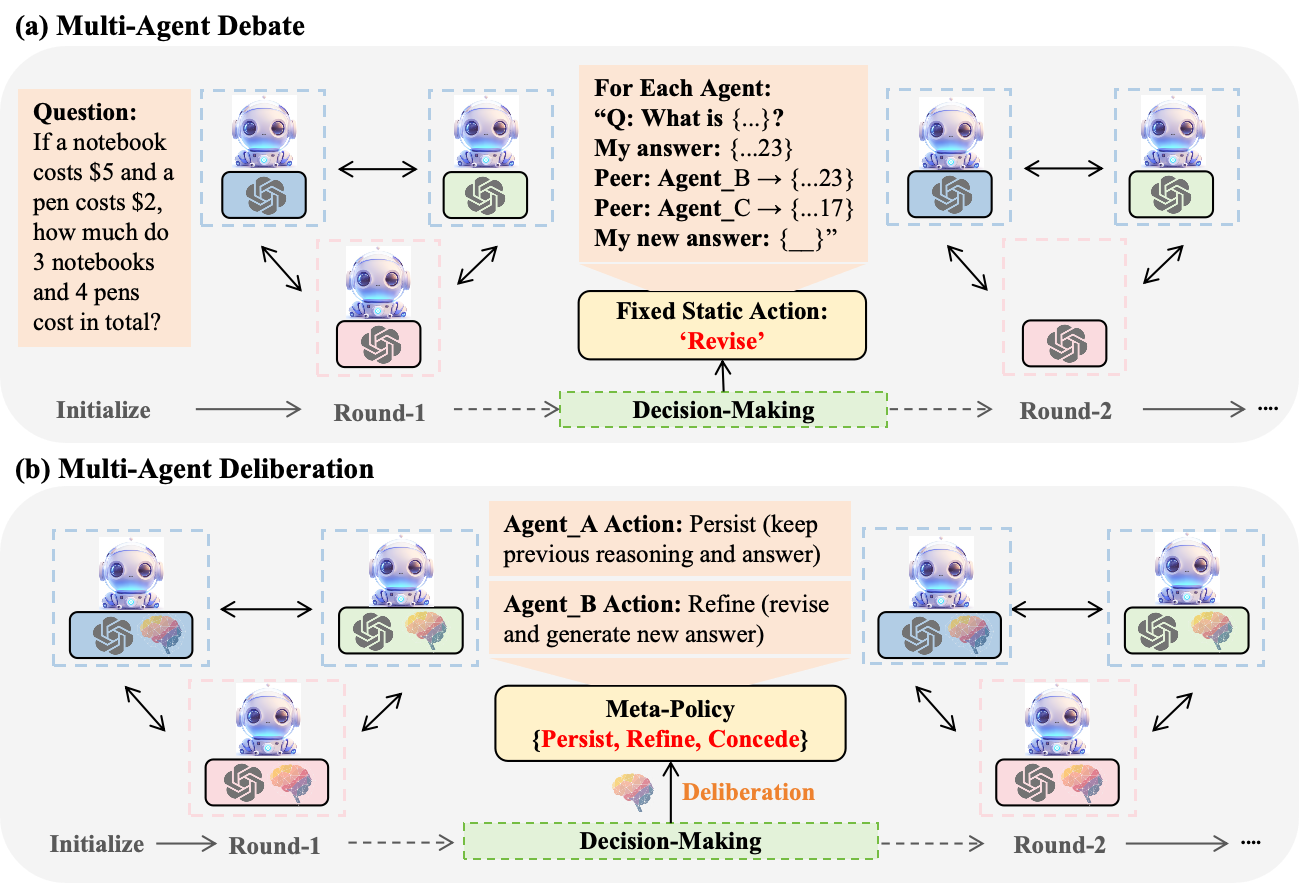}
\caption{Evolving from fixed static debate to trainable meta-policy deliberation, with dynamic strategy adaptation for multi-agent collaboration.}
\label{fig:fig_teaser}
\end{figure}

Recent efforts to overcome the static protocol bottleneck have largely focused on dynamic, system-level optimizations \cite{zhao2023competeai}, which we categorize as \textit{macro-level orchestration}. 
These methods 
optimize the connections between agents, but critically neglect the deliberative \textit{meta-cognition} \cite{sun2006modeling,yeung2012metacognition,koriat2015metacognition} within each agent.
This \textit{Meta-Cognitive Blindspot} prevents agents from differentiating 
between confident judgments and uncertain hypotheses, causing them to uniformly apply external protocols irrespective of their internal cognitive state, falling back to a
\textit{Static Deliberation Strategy}.
That is, agents employ fixed or pre-programmed response behaviors, lacking a strategic faculty to adaptively choose among persisting, refining, or conceding actions based on context and internal confidence. 

To address these deficiencies, we introduce the \textbf{Meta-Policy Deliberation Framework (MPDF)}, which formulates multi-agent collaboration as a decentralized partially observable Markov decision process (Dec-POMDP). Unlike prior approaches relying on static heuristics or predefined rules, our framework equips each agent with a lightweight, decoupled policy network. Each agent explicitly reasons about its internal cognitive state and strategically selects from discrete deliberation acts such as \textsc{Persist}, \textsc{Refine}, or \textsc{Concede}. This meta-cognitive capability enables agents to dynamically calibrate their behavior according to evolving confidence and situational context. 

To robustly optimize this policy under sparse and noisy team-level feedback, we introduce \textbf{SoftRankPO}, a scale-resilient reinforcement learning algorithm. SoftRankPO helps stabilize policy gradients by converting raw rewards into rank-based advantages derived from smooth Gaussian percentiles, thereby mitigating reward-scale sensitivity and high-variance challenges.

Our main contributions are:
\begin{itemize}[nosep,noitemsep]
    \item A meta-policy framework formulating multi-agent LLM reasoning as a Dec-POMDP for adaptive coordination through learned decision-making dynamics.

    \item SoftRankPO, a scale-resilient policy optimization algorithm that stabilizes learning via smooth rank-based advantage shaping, enabling low-variance and reliable convergence across diverse reward regimes.

    \item Consistent accuracy gains over competitive baselines on six reasoning benchmarks by combining meta-cognitive coordination and scale-resilient optimization.
\end{itemize}

\section{Related Works}

\paragraph{Multi-Agent LLM Collaboration.}

Beyond large language models \cite{chen2025tourrank,ping2025hdlcore,li2025climatellm}, collaborative multi-agent LLM frameworks for reasoning and planning 
\cite{motwani2024malt,ishibashi2024self,trirat2024automl,yan2025beyond,dai2025multi,chen2024comm,chang2025survey,zhang2025evoflow}
mainly follow two paradigms.
First, prestructured approaches rely on fixed coordination schemes \cite{liu2023dynamic}, like debate \cite{du2023improving,liu2024groupdebate} and peer review pipelines.
These systems adopt static topologies such as chains, trees, or graphs to guide communication \cite{du2023improving,qian2024scaling}. 
Second, self-organizing methods adapt interactions during inference by restructuring collaboration graphs using Monte Carlo tree search, evolutionary updates, and pruning \cite{hu2024automated,shang2024agentsquare,zhang2024cut}. 
Some 
frame workflow design as program synthesis over modules \cite{zhuge2024gptswarm,zhang2024aflow}, or assign dynamic roles 
across agents \cite{hu2024routerbench,yue2025masrouter}.
These methods have advanced the design space of LLM-based collaboration, but rely on heuristic control or static policies \cite{wei2025lero}. 
We overcome this limitation by introducing 
closed-loop learning to adapt collaboration based on downstream utility.

\paragraph{Multi-Agent Reinforcement Learning for LLMs.}

Reinforcement learning (RL) \cite{chen2022ptde,chen2022commander,hu2024review} for multi-agent LLM systems aim to train coordinated, role-specialized agents beyond static prompting or supervised tuning \cite{zheng2023progressive,zhu2025lamarl,peiyuan2024agile,park2025maporl,liao2025marft,han2025joyagents,wan2025rema,zhuang2024yolo}. 
For example, MAPoRL \cite{park2025maporl} co-trains multiple LLMs through multi-turn debates using verifier-shaped collaborative rewards, and LERO \cite{wei2025lero} employs LLMs in an evolutionary MARL loop to generate hybrid reward and observation functions. 
To support scalable optimization, frameworks like Maestro \cite{yang2025maestro} and MHGPO \cite{chen2025heterogeneous} adopt centralized coordination with decentralized policy learning.
Others, such as ReSo \cite{zhou2025reso} and Verco \cite{li2024verco}
introduce dynamic agent selection and interpretable communication protocols.
Meanwhile, ML-Agent \cite{liu2025ml} and RAGEN \cite{wang2025ragen} explore step-wise RL and long-horizon reasoning, showing that LLM agents can self-improve across complex decision trajectories. 
Together, these works represent a growing trend toward training LLM-based MAS that learn to communicate and adapt under reinforcement objectives \cite{yang5819182toward,zhang2024multi,sun2024llm,chen2025discriminative}.

\begin{figure*}[t]
\centering
\includegraphics[width=\linewidth]{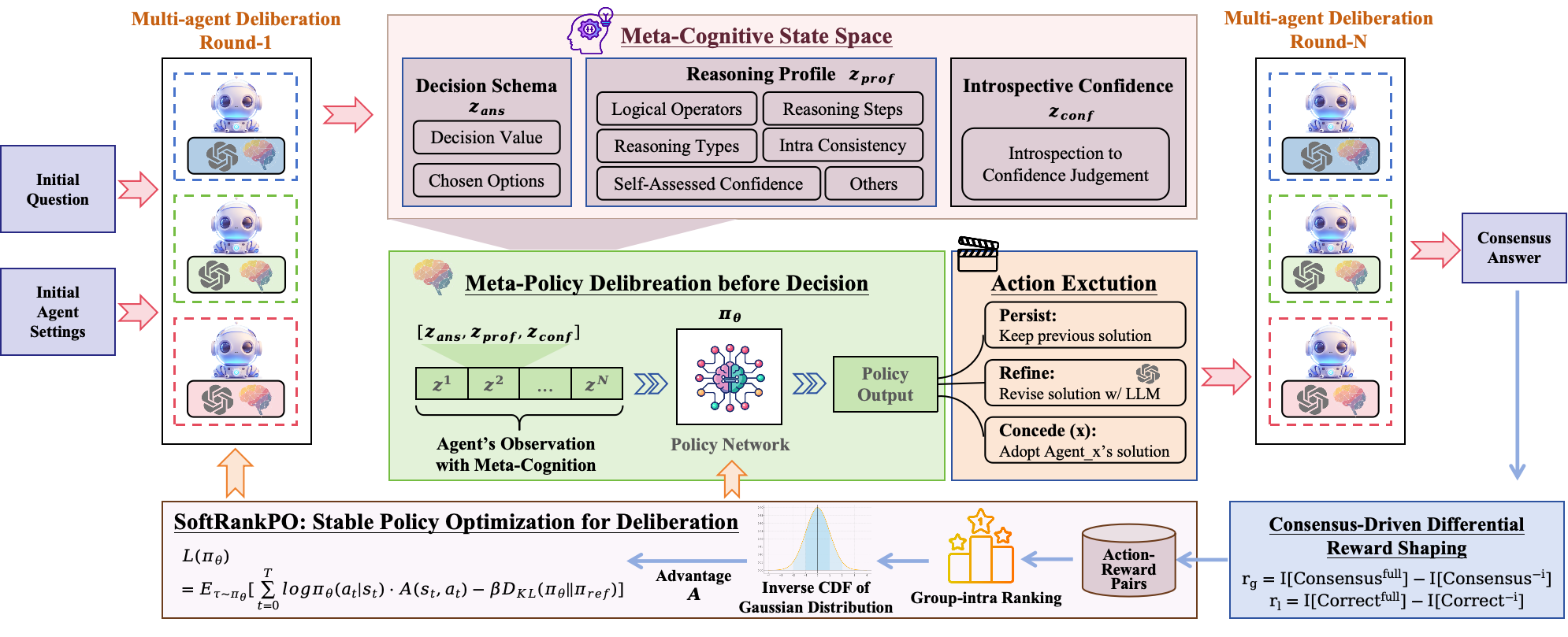}
\caption{An overview of the Meta-Policy Deliberation Framework (MPDF). (a) Meta-Cognitive State Representation: Each agent constructs a cognitive state with decision schema, reasoning profile and introspective confidence. (b) Meta-Policy Deliberation: The agent's meta-policy network selects one of three high-level actions: Persist (if confident), Refine (to improve its own solution), or Concede (to yield to a peer solution). (c) Environment Interaction: The chosen action leads to a new response, yielding a reward and transitioning the system to the next state. (d) SoftRankPO: a stable policy optimization RL algorithm.}
\label{fig:fig_main}
\end{figure*}

\section{Method}

We formalize the multi-agent deliberation process as a decentralized partially observable Markov decision process (Dec-POMDP), defined by the tuple $\langle \mathcal{I}, \mathcal{S}, \mathcal{A}, \mathcal{T}, \mathcal{R}, \Omega, \mathcal{O}, \gamma \rangle$. Here, $\mathcal{I}$ is the set of $N$ agents, $\mathcal{S}$ is the unobserved global state space representing the true problem-solving state, and $\mathcal{A} = \times_{i \in \mathcal{I}} \mathcal{A}^i$ is the joint action space composed of individual agent actions from $\mathcal{A}^i$.

At each deliberation round $t \in \{0, \dots, T-1\}$, each agent $i \in \mathcal{I}$ receives a local observation $o_t^i \in \Omega$ and executes a deliberative action $a_t^i \in \mathcal{A}^i$ according to its policy $\pi^i_\theta(a_t^i | o_t^i)$. The environment transitions based on the function $\mathcal{T}(s_{t+1} | s_t, \mathbf{a}_t)$, where $\mathbf{a}_t$ is the joint action. Subsequently, each agent receives a reward $r_t^i = \mathcal{R}^i(s_t, \mathbf{a}_t)$, shaped to reflect its marginal contribution to the team's objective. The collective goal is to learn an optimal, decentralized policy $\pi^* = (\pi^{1*}, \dots, \pi^{N*})$ that maximizes the expected discounted return:
\begin{equation}
\pi^* = \arg\max_{\pi} \ \mathbb{E}_{\pi} \left[ \sum_{t=0}^T \gamma^t \sum_{i \in \mathcal{I}} r_t^i \right]
\end{equation}
The primary research challenge, and the core of our contribution, lies in the specific instantiation of the observation space $\Omega$ and the action space $\mathcal{A}$. Traditional approaches often rely on high-dimensional, unstructured text, leading to sample inefficiency and a focus on superficial linguistic patterns. In contrast, we propose a novel \textbf{Meta-Cognitive Deliberation Framework}, detailed below, which defines these spaces at a higher level of strategic abstraction.


\subsection{The Meta-Cognitive Deliberation Framework}
\label{sec:framework}

Our framework is designed to overcome the \textit{Meta-Cognitive Blindspot} and \textit{Static Deliberation Strategy} shortcomings of existing multi-LLM-agent frameworks by equipping each agent with the ability to reason about cognitive state and to select from high-level strategic acts.

\subsubsection{The Meta-Cognitive State Space}
We define an agent's local state $z_t^i$ 
on a structured, low-dimensional vector summarizing its meta-cognitive status. This design choice forces the policy to learn abstract strategic reasoning rather than superficial text matching. The state $z_t^i$ is composed of three key components, with full feature specifications and prompt templates detailed in Appendix~B.

\noindent\textbf{Decision Schema ($z_{ans}$)} represents the agent's conclusive output, providing a concise summary of \textit{what} the agent believes. It is extracted directly from the agent's textual response by parsing the final output into a standardized format (e.g., a scalar value for mathematical problems or a one-hot vector for multiple-choice questions).

\noindent\textbf{Reasoning Profile ($z_{prof}$)} is a vector of self-reported features that summarizes \textit{how} the agent reached its belief. These features are produced by the LLM agent in a structured format as part of its single generative step, capturing attributes like the number of reasoning steps, the logical operators employed (\textit{e.g.}, algebraic, arithmetic), and a self-assessed confidence score. This co-generation process yields meta-data without incurring additional inference cost.

\noindent\textbf{Introspective Confidence ($z_{conf}$)} 
is generated via a dedicated \textit{introspection step}, where a critic model, specifically the same LLM backbone acting in a evaluation capacity, is prompted to assess the correctness of the agent's full reasoning trace given the original problem. The critic's judgment, which may take the form of a graded assessment or confidence-aware verdict (e.g., leaning toward ``Correct'' or ``Incorrect'' with varying strength), is then embedded into a vector representation, providing a signal for deliberation.

An agent's full observation $o_t^i$ for decision-making is composed of its own local meta-cognitive state and those of its peers: $o_t^i = (z_t^i, \{z_t^j\}_{j \neq i})$. The \textbf{policy network}, as a core component of our deliberation framework, then utilizes a cross-attention mechanism to integrate its introspective assessment with the social context provided by its peers. Please see Appendix B for the full policy network structure.

\subsubsection{The Deliberative Action Space}
To address the issue of static strategies, we move beyond simple communicative acts to a set of high-level, strategic deliberative acts, $\mathcal{A}^i$. These actions represent fundamental choices in a collaborative reasoning process, allowing the agent to learn a truly adaptive policy. The action space is defined as:
\textbf{Persist}, deciding to maintain and defend one's current solution against competing alternatives;
\textbf{Refine}, an act of triggering an internal process of self-correction and iterative improvement; and
\textbf{Concede}, a strategic deference to a peer's solution based on an assessment of its higher quality.

By learning a policy $\pi^i_\theta(a_t^i | o_t^i)$ over this action space, our agents develop a dynamic, context-dependent deliberation strategy. They learn not just how to solve the problem, but how to \textbf{behave optimally as a member of a reasoning collective}---deciding when to lead, when to follow, and when to rethink, 
unlocking more efficient and robust reasoning.

\subsection{SoftRankPO: A Stable Policy Optimization Algorithm for Deliberation}

Learning a high-level deliberation policy $\pi_\theta$ within our Dec-POMDP setting is challenging because the reward signal is sparse and highly non-uniform across batches. The \textsc{Refine} action in particular depends on stochastic LLM generations, leading to heavy-tailed returns and rounds where almost all agents make similar choices. In these cases the within-batch standard deviation can be extremely small, while the across-batch variance remains large. Moreover, the step-wise differential gains and leave-one-out consensus rewards can amplify this effect: high-consensus rounds often exhibit near-zero within-batch variance, whereas batches with more diverse or difficult candidates induce much larger variability. Standard policy gradients, including PPO and GRPO, construct advantages directly from raw reward magnitudes, so their gradient variance is sensitive to such scale and variance heterogeneity even after per-group normalization. In regimes with near-degenerate batches, per-group standard-deviation normalization can become numerically brittle.

In practice, the effective reward scale can differ substantially across tasks and batches, and even simple affine rescaling $R \mapsto \alpha R + \beta$ can significantly inflate or shrink gradient magnitudes. Since the gradient variance satisfies $\operatorname{Var}[\nabla \mathcal{L}] \propto \operatorname{Var}[R]$, learning can become unstable under batch-specific reward shifts. Although techniques such as baseline subtraction, batch normalization, or group-wise standardization help reduce variance, they do not remove the underlying dependence on reward scale, which remains a fragile foundation in collaborative systems where agents operate with heterogeneous criteria and subjective feedback.

\subsubsection{SoftRank: Placing \emph{Order} before \emph{Scale}}
\label{sec:method:softrank}

We introduce SoftRankPO, a KL-regularized policy-optimization method that aligns
a trainable policy with \emph{ranked} preferences. By divorcing learning signals from raw reward magnitudes, SoftRankPO reduces sensitivity to reward scale and variance while still operating inside a familiar KL trust region.

Consider a state \(s\) with reward vector
\(\mathbf R=(R_1,\dots,R_K)\) for the \(K\) candidate actions.
We first convert the ordinal information into a
standardized score through the mapping
\begin{equation}
p_i=\Bigl(\frac{\operatorname{rank}(R_i)+\tfrac12}{K}\Bigr)^{\tau}, 
\qquad 
A_i=\frac{A_i^\star-\bar A^\star}
          {\mathrm{std}(\mathbf A^\star)+\varepsilon},
\label{eq:softrank_long}
\end{equation}
where \(\Phi^{-1}\) denotes the inverse standard normal CDF of $\mathcal N(0,1)$,
$A_i^\star=\Phi^{-1}(p_i)$ and $\bar A^\star=\tfrac1K\sum_{j}A_j^\star$. The temperature parameter $\tau > 0$ controls the contrast between high- and low-ranked actions, enabling smoother or sharper advantage shaping. Equation~\eqref{eq:softrank_long} preserves the relative ranking of rewards, is invariant to affine transformations of
\(\mathbf R\), and produces approximately zero-mean, bounded-variance scores whenever
\(K\) is reasonably large. In Appendix~A we formalize these properties and show that the resulting advantages are sub-Gaussian, providing a stable and well-behaved signal for policy optimization.

\subsubsection{KL-constrained Rank-matching Objective}
\label{sec:method:objective}

Let \(\pi_{\mathrm{ref}}\) be a frozen reference policy obtained, for
example, from supervised fine-tuning.
Under SoftRankPO, we optimize the KL-regularized objective
\begin{equation}
\begin{split}
\max_{\pi_\theta}
\; \mathbb{E}_{s\sim\mathcal{D}}\Bigl[
\mathbf{A}(s)^{\!\top}\log\pi_\theta(\cdot\mid s)
- &\beta\,\mathrm{KL}\bigl(\pi_\theta\;\|\;\pi_{\mathrm{ref}}\bigr)\\
&\quad\;+\;\lambda\,\mathcal{H}\bigl(\pi_\theta\bigr)
\Bigr],
\end{split}
\label{eq:soft_obj_split}
\end{equation}
where \(\beta>0\) controls the KL trust region and
\(\mathcal H\) denotes Shannon entropy. Writing this in terms of the implicit KL-induced reward
\(R_\theta(s,a)=\beta\log\frac{\pi_\theta(a\mid s)}{\pi_{\mathrm{ref}}(a\mid s)}\)
and using the zero-mean property \(\sum_{i}A_i(s)=0\), we show in Appendix~A that maximizing
\(J(\theta)\) is equivalent (up to constants) to minimizing the following \textbf{rank-matching loss}:
\begin{equation}
\mathcal L_{\mathrm{SR}}(\theta)=
\mathbb E_{s}\!\Bigl[\frac1K\sum_{i=1}^{K}
\bigl(\,R_\theta(s,a_i)-\bar R_\theta(s)-R(s,a_i)+\bar R(s)\bigr)^{\!2}
\Bigr],
\label{eq:rank_mse}
\end{equation}
where \(\bar R(\cdot)\) and \(\bar R_\theta(\cdot)\) denote within-state averages. Hence SoftRankPO learns logits whose \emph{relative shape} mirrors that
of the true ranked rewards, without requiring explicit partition functions. It can be interpreted
either as a KL-regularized policy optimizer or as a rank-regression procedure that aligns implicit rewards with task rewards.

\paragraph{Optimization behaviour and theoretical properties.}
\label{sec:method:theory_summary}
The SoftRank advantages are centered and have bounded variance, which implies that the partition-function term cancels in the gradient and the update admits a clean analytical form. In Appendix~A we derive the gradient of $\mathcal L_{\mathrm{SR}}(\theta)$, show how it decomposes into advantage-weighted learning, entropy-variance control, and KL trust-region coupling, and restate a standard result characterizing the KL-constrained optimum. We also provide a variance comparison with GRPO and a first-order convergence rate under common smoothness assumptions. 

\subsection{Consensus-Driven Differential Reward Shaping}

To enable fine-grained credit assignment in decentralized deliberation, we design a reward framework grounded in the principle of marginal contribution. Instead of assigning scalar rewards directly based on outcome quality, we decompose the reward signal into two complementary components: local self-improvement and global influence on consensus. 

The \emph{step-wise delta reward} quantifies an agent’s immediate self-improvement, defined as the change in correctness of its own output before and after an action: $r_{\text{local}} = \mathbb{I}[\text{Correct}^{\text{after}}] - \mathbb{I}[\text{Correct}^{\text{before}}]$,
encouraging agents to increase their individual answer quality. This step-wise shaping is conceptually related to prior MARL work on social deduction games, where agents are rewarded for improving others’ answers~\cite{sarkar2025training}, but here we focus on self-improvement and combine it with a consensus-based marginal term.

The \emph{lookahead consensus reward} captures an agent’s contribution to the final consensus, defined by its marginal impact on the outcome. Let $\text{Consensus}^{\text{full}}$ and $\text{Consensus}^{-i}$ denote the correctness of the final group decision with and without agent $i$’s input. The global reward is computed as $r_{\text{global}} = \mathbb{I}[\text{Consensus}^{\text{full}}] - \mathbb{I}[\text{Consensus}^{-i}]$,
and in practice $\text{Consensus}^{-i}$ is recomputed from the cached responses in the same round without requiring additional rollouts.

Together, the total reward $r = r_{\text{local}} + r_{\text{global}}$ provides structured, causally grounded supervision that aligns individual optimization with collective utility, enabling stable and targeted training of multi-agent deliberation policies.

\subsection{Supervised Bootstrapping and MARL Fine-Tuning}

Our training pipeline follows a two-stage process that integrates supervised guidance with variance-robust reinforcement learning, enabling stable convergence of the deliberation strategy. First, for robust initialization, we pre-train the policy network $\pi_{\theta}$ on an expert dataset $\mathcal{D}_{\text{sft}}=\{(s, a^{\star})\}$, where labels $a^{\star}$ are derived from oracle solutions, by minimizing a standard cross-entropy loss. This yields a supervised policy $\pi_{\text{sft}}$, which we also use as the fixed reference policy in the KL regularization, i.e., $\pi_{\text{ref}} = \pi_{\text{sft}}$ in Eq.~\ref{eq:soft_obj_split}. We initialize $\pi_{\theta}$ from $\pi_{\text{sft}}$ before reinforcement learning. Subsequently, we generate a static offline corpus $\mathcal{D}_{\text{off}}$ by executing rollouts with $\pi_{\text{sft}}$. This corpus stores tuples of encountered meta-cognitive states and the resulting group-wise reward vectors, i.e., $\mathcal{D}_{\text{off}} = \{(s, \mathbf{R})\}$. In the second stage, we fine-tune $\pi_{\theta}$ on $\mathcal{D}_{\text{off}}$ using our SoftRankPO objective (Eq.~\ref{eq:soft_obj_split}), aligning the policy with the stable, rank-based advantages derived from the offline data. The KL-divergence term ensures that the updated policy remains within a trust region of the sensible SFT initialization.

\section{Experiment}


We evaluate our framework on three standard reasoning tasks for LLM agents \cite{hendrycks2020measuring,cobbe2021training,hendrycks2021measuring,chen2021evaluating,du2023improving}.
\emph{(i) Mathematical reasoning} is assessed on \textbf{GSM8K}, \textbf{MATH}, \textbf{AIME}, and \textbf{AMC}.
\emph{(ii) General reasoning} is measured with the \textbf{MMLU} benchmark.
\emph{(iii) Code generation} is evaluated on \textbf{HumanEval}.
For all data sets, we follow the original splits and report the official metrics: \emph{Solve Rate} for math datasets, \emph{Accuracy} in MMLU and \emph{Pass@1} in HumanEval.

\paragraph{Baselines.}
We compare MPDF against a comprehensive suite of baselines, including single-agent methods: \textbf{Vanilla}, \textbf{CoT} and \textbf{SC}; multi-agent sequential and debate frameworks: \textbf{PHP} and \textbf{LLM-Debate}; and dynamic workflow systems: \textbf{DyLan}, \textbf{AgentPrune}, \textbf{AFlow} and \textbf{GPTSwarm}. 

\paragraph{Implementation Details.}
Following common practice, we set the number of agents and rounds to 3 by default. We instantiate agents with \textbf{Llama-3.1-8B-Instruct}, \textbf{Llama-3.2-3B-Instruct}, \textbf{Qwen2.5-7B-Instruct} and \textbf{Qwen2.5-3B-Instruct}. For all results, we report the average over 3 random seeds. Detailed hyperparameters and all settings are provided in Appendix C to ensure full reproducibility.

\begin{table*}[t]
\small
\centering
\rowcolors{1}{gray!10}{white}
\begin{tabular}{l|cc|cccccc|c}
\toprule
\textbf{Model} & \textbf{MA} & \textbf{Type} & \textbf{GSM8K} & \textbf{MATH} & \textbf{AIME} & \textbf{AMC} & \textbf{MMLU} & \textbf{HumanEval} & \textbf{Avg} \\
\midrule
Vanilla         & \xmark & Single        & 72.71 & 66.91 & 3.33  &  8.43  & 57.83 & 48.17          & 42.90 \\
CoT             & \xmark & Single        & 74.00 & 68.18 & 3.33  & 12.05  & 61.60 & 51.83          & 45.17 \\
SC              & \xmark & Single        & 80.21 & 73.09 & 4.44  & 12.05  & 68.64 & 56.10          & 49.09 \\
\hline
PHP             & \cmark & Sequential        & 80.36 & 74.36 & 4.44  & 15.67  & 68.72 & 56.70          & 50.04 \\
LLM-Debate      & \cmark & Debate        & 83.47 & 78.91 & 5.56  & 18.07  & 67.41 & 58.54          & 51.99 \\
DyLAN           & \cmark & Dynamic        & 82.63 & 76.55 & 6.67  & 19.28  & 66.84 & 60.37          & 51.50 \\
GPTSwarm        & \cmark & Dynamic       & 84.07 & 73.64 & 5.56  & 15.66  & 69.48 & 57.32          & 51.14 \\
AgentPrune      & \cmark & Dynamic       & 84.53 & 79.45 & 4.44  & 14.46  & 68.82 & 57.92          & 51.60 \\
AFlow           & \cmark & Dynamic       & 83.40 & 75.63 & 3.33  & 10.84  & 69.07 & 62.16 & 50.74 \\
\hline
MPDF: SFT             & \cmark & Deliberation & 83.55 & 79.27 & 5.56  & 16.87  & 68.73 & 58.54 & 52.09 \\
MPDF: SFT + PPO       & \cmark & Deliberation & 82.79 & 79.82 & 4.44  & 13.25  & 68.15 & 56.62 & 50.85 \\
MPDF: SFT + GRPO      & \cmark & Deliberation & 85.52 & 81.45 & 7.78  & 20.48  & 70.22 & 60.98 & 54.40 \\
MPDF: SFT + SoftRankPO& \cmark & Deliberation & \textbf{86.35} & \textbf{82.72} & \textbf{7.78} & \textbf{22.89} & \textbf{70.87} & \textbf{63.82} & \textbf{55.74} \\
\bottomrule
\end{tabular}
\caption{Comparison of performance across reasoning tasks. “MA” denotes the use of multiple agents (\cmark) versus single-agent baselines (\xmark). “Avg” reports the average accuracy across all applicable benchmarks. Our method consistently outperforms both single-agent and multi-agent baselines.}
\label{tab:main_result}
\end{table*}

\begin{table}[t]
\small
\centering
\rowcolors{1}{gray!10}{white} 
\begin{tabular}{l|cccc}
\toprule
\textbf{Model} & \textbf{LMA-8B} & \textbf{LMA-3B} & \textbf{Qwen-7B} & \textbf{Qwen-3B} \\
\midrule
Vanilla      & 72.71 & 38.64 & 90.90 & 83.62 \\
CoT          & 74.00 & 50.49 & 91.05 & 84.23 \\
SC           & 80.21 & 54.51 & 93.33 & 88.17 \\
\hline
PHP          & 80.36 & 62.47 & 92.87 & 86.81 \\
LLM-Debate   & 83.47 & 76.19 & 93.56 & 87.34 \\
DyLAN        & 82.63 & 76.65 & 93.25 & 87.79 \\
GPTSwarm     & 84.07 & 68.84 & 92.49 & 86.50 \\
AgentPrune   & 84.53 & 65.35 & 92.12 & 86.35 \\
AFlow        & 83.40 & 59.67 & 90.22 & 82.11 \\
\hline
MPDF         & \textbf{86.35} & \textbf{80.52} & \textbf{94.47} & \textbf{89.77} \\
\bottomrule
\end{tabular}
\caption{Performance on GSM8K across different LLM backbones. Results demonstrate consistent performance gains across model scales and pre-training settings.}
\label{tab:backbone_comparison}
\end{table}

\begin{table}[t]
\small
\centering
\rowcolors{1}{gray!10}{white} 
\begin{tabular}{l|rrr}
\toprule
\textbf{Dataset} & \textbf{LLM-Debate} & \textbf{MPDF} & \textbf{Improvement (\%)}\\
\midrule
GSM8K      &    8,608     &    6,154     & \textbf{28.53\%} \\
MATH       &    7,474     &    4,346     & \textbf{41.85\%} \\
AIME       &   19,321     &   14,077     & \textbf{27.15\%} \\
AMC        &   17,917     &   12,620     & \textbf{29.58\%} \\
MMLU       &   10,908     &    7,322     & \textbf{32.87\%} \\
HumanEval  &    3,054     &    2,616     & \textbf{14.34\%} \\
\midrule
\textbf{Total} &  67,282   &   47,135     & \textbf{29.96\%} \\
\bottomrule
\end{tabular}
\caption{Average total token count (input and output combined) per sample across multiple datasets.}
\label{tab:token_cost_small}
\end{table}

\subsection{Main Results}

\paragraph{Our Multi-Agent Deliberation Framework Outperforms Baselines.}
As shown in Table~\ref{tab:main_result}, our deliberation-based multi-agent framework achieves the highest average accuracy (55.74\%) across six diverse reasoning benchmarks, outperforming both debate-style methods (e.g., DyLAN 51.50\%) and dynamic approaches (e.g., AgentPrune 51.60\%). Unlike single-agent methods (e.g., Vanilla, CoT, SC), which lack inter-agent coordination, and debate-style models, which suffer from unstable decision-making and weak convergence, our framework jointly optimizes decision-making and consensus formation to enhance multi-agent reasoning. While routing methods attempt to improve efficiency through static interaction graphs and heuristic specialization, they remain limited by their inability to support dynamic deliberation or mutual influence during execution. In contrast, our approach establishes a unified, meta-cognitive deliberation process that enables agents to collaborate iteratively toward shared goals, improving accuracy especially in high-complexity tasks such as AMC and AIME. 


\begin{figure}[t]
\centering
\includegraphics[width=\columnwidth]{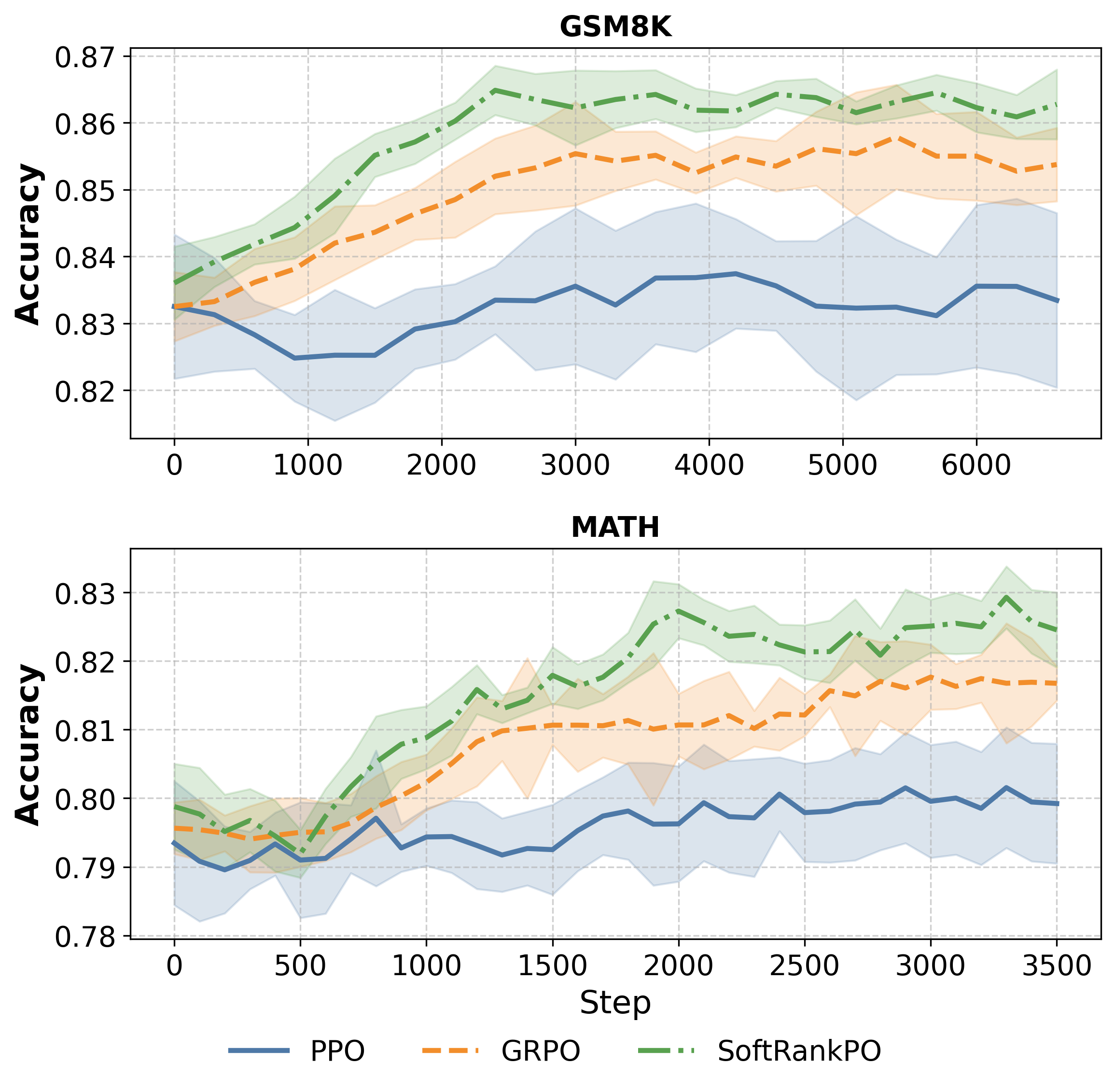}
\caption{Accuracy curves during deliberative training on GSM8K (top) and MATH (bottom) using the LLaMA3-8B backbone. Our method consistently achieves stable and faster convergence across both benchmarks.}
\label{fig:fig_acc_curve}
\end{figure}

\paragraph{Consistent Gains Across Diverse LLM Backbones.}
To verify that our framework generalizes beyond a single model family, we conduct additional experiments using LLaMA3 and Qwen2.5 as representative backbones. Table~\ref{tab:backbone_comparison} summarizes the results, demonstrating consistent and robust accuracy gains across all settings. On LMA-3B, the weakest backbone in our evaluation, accuracy improves from 68.84\% (GPTSwarm) to 80.52\%, indicating that the learned meta-policy effectively compensates for limited model capacity. Scaling up to LMA-8B still yields a positive margin, lifting the best baseline from 84.53\% to 86.35\%. Switching to Qwen models is particularly informative, as these checkpoints already benefit from extensive instruction tuning; Qwen-7B alone reaches 91.05\%, yet deliberative training boosts it further to 94.47\%; and Qwen-3B accuracy increases from 84.23\% to 89.77\%. 

\paragraph{Lower Token Cost with Competitive Effectiveness.}
To facilitate comparison, we report results against the widely adopted LLM-Debate baseline.
Our method achieves a substantial reduction in average token use (47.1k), measured as both input and output tokens, outperforming LLM-Debate (67.2k) while maintaining or surpassing its accuracy (Table~\ref{tab:token_cost_small}). LLM-Debate operates by aggregating all agent responses and prompting the model to synthesize an answer in a single pass, but lacks a mechanism for adaptive filtering. As a result, it often generates redundant and unfocused interactions, inflating token cost without proportional gains in solution quality. In contrast, our method introduces meta-cognitive policies that guide agents to act reflectively. Through reinforcement learning over interaction trajectories, agents learn to retain high-confidence outputs and intervene only when revisions are likely to yield substantial gains. This behavioral shift is evident in the increased \texttt{PERSIST} action rate (78.8\% vs. 19.1\% in baseline SFT), as shown in Figure~\ref{fig:action_policy}, a move toward confident retention.

\subsection{Ablation and Analysis Studies}

\paragraph{From Fine-Tuning to SoftRankPO: Multi-Agent RL for Deliberation.}

Our framework transitions from supervised fine-tuning (SFT) to progressively more effective reinforcement learning (RL) strategies that enhance agent coordination. As shown in Table~\ref{tab:main_result}, even the SFT-only variant performs competitively (52.09\%), highlighting the strength of deliberation-based reasoning. However, it lacks outcome-driven feedback, limiting its ability to refine decision-making. Vanilla PPO performs worse (50.85\%) due to unstable credit assignment and poor convergence in our multi-agent setting. As shown in Figure~\ref{fig:fig_acc_curve}, PPO struggles especially on harder tasks (e.g., -3.62\% on AMC), where its value-based updates fail to provide stable guidance. GRPO mitigates this by using group-relative ranking, stabilizing updates through peer comparison and improving performance on tasks like GSM8K (+1.97\%) and MATH (+2.18\%), but its reliance on raw reward magnitudes leads to sensitivity under noisy or heavy-tailed feedback. 

\paragraph{Communication Rounds and Number of Agents.}

\begin{figure}[t]
\centering
\includegraphics[width=\columnwidth]{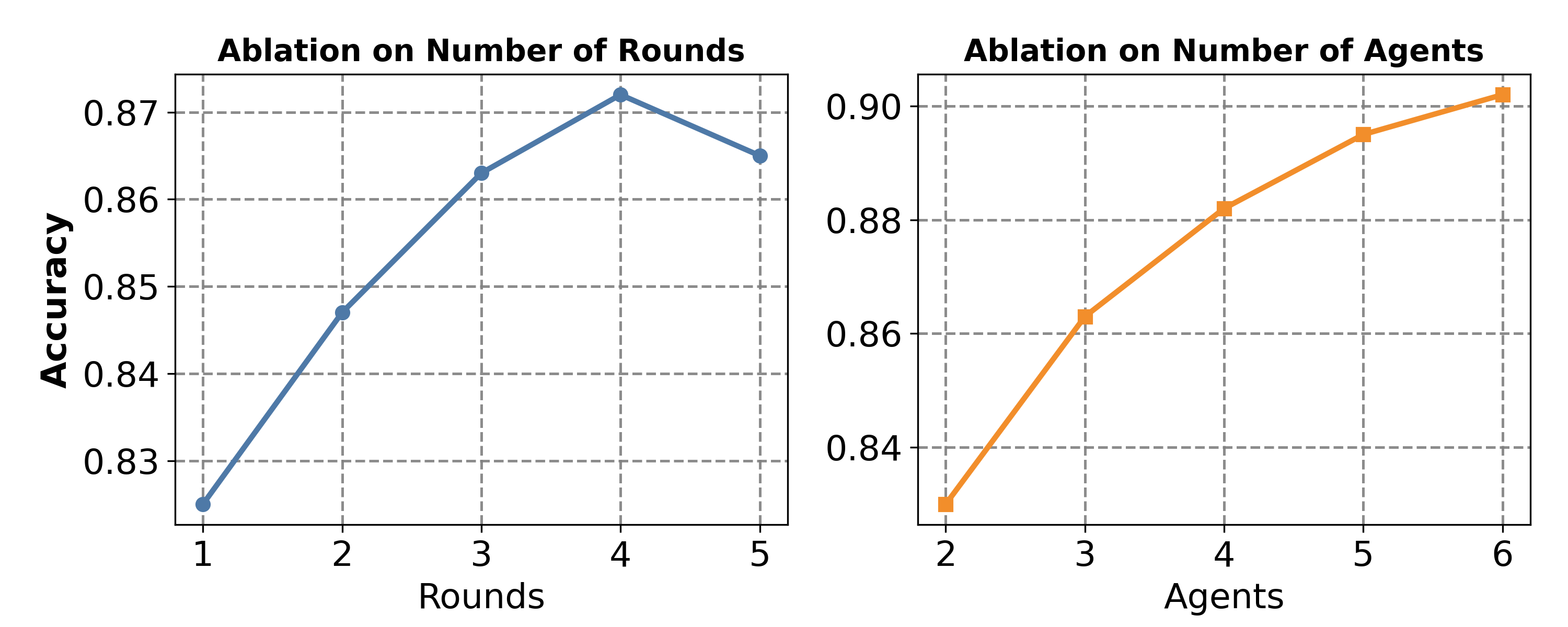}
\caption{Our framework scales well with more rounds and agents, with diminishing returns beyond optimal settings.}
\label{fig:fig_ab_round_agent}
\end{figure}

Figure~\ref{fig:fig_ab_round_agent} shows how communication rounds ($M$) and agent number ($N$) affect final performance. Increasing $M$ from 1 to 4 steadily improves accuracy, validating the importance of multi-turn deliberation. However, performance slightly drops at 5 rounds, suggesting diminishing returns and possible noise accumulation, highlighting the need to calibrate deliberation depth. As $N$ increases from 2 to 6, accuracy consistently improves.
Unlike unstructured setups that suffer from conflict or stagnation, our reflective coordination continues to benefit from additional agents. 

\begin{figure}[t]
\centering
\includegraphics[width=1.05\columnwidth]{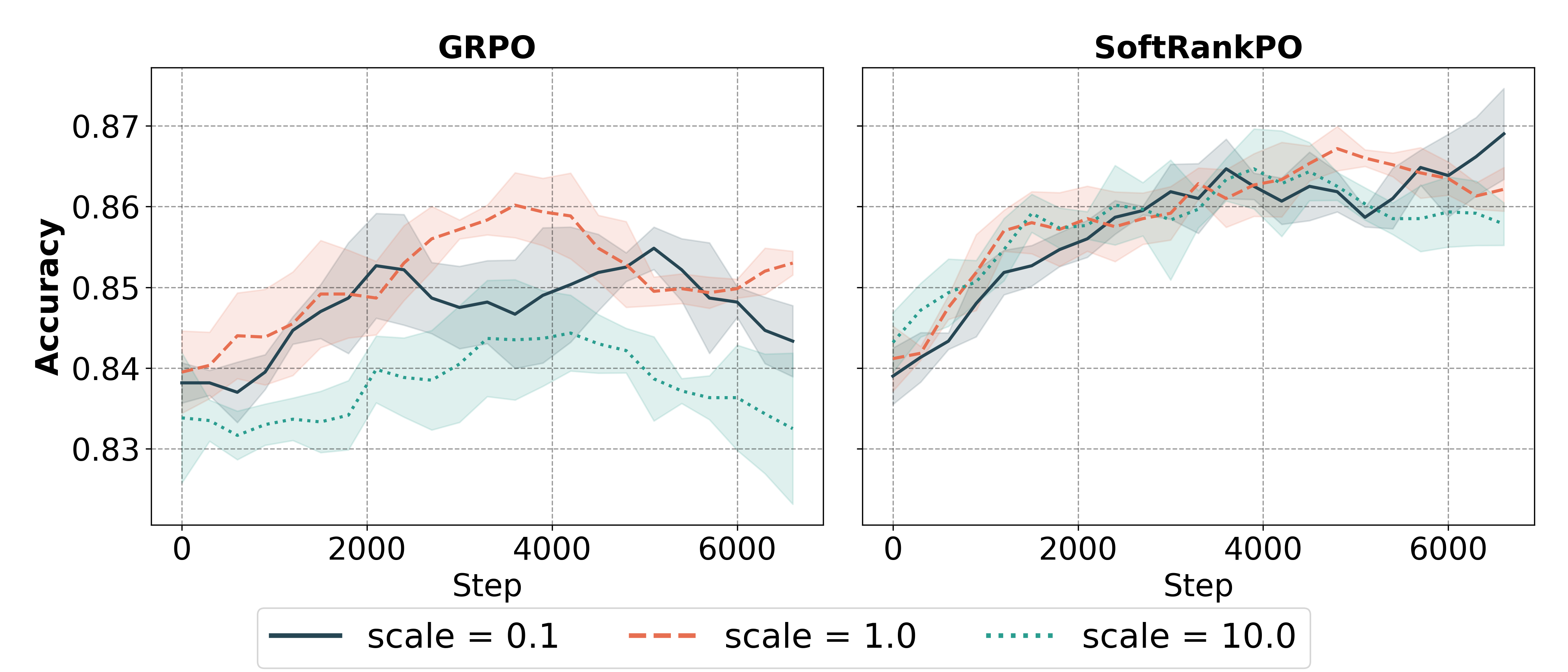}
\caption{Stability under different exponential reward scales. SoftRankPO remains robustness across multiple scales.}
\label{fig:reward_scale}
\end{figure}

\paragraph{Robustness to Reward Scale.}
We conduct an ablation where the reward signal is rescaled by exponential factors $\{0.1,1.0,10.0\}$. As shown in Figure~\ref{fig:reward_scale}, GRPO suffers from noticeable instability: different scales induce diverging learning curves, and all variants exhibit a pronounced drop in performance after early gains. This behavior stems from GRPO's reliance on perstate $z$-score advantages, where an outlier can dominate the local mean and variance, 
leading to unstable gradient updates. In contrast, SoftRankPO is robust across reward scales. Its accuracy consistently improves and stabilizes around the same level regardless of the reward scaling, with visibly smoother and more stable curves. This empirical observation aligns with our theoretical analysis. By inducing a zero-mean, bounded-variance advantage shaped by standard Gaussian, SoftRankPO effectively decouples learning dynamics from the scale of rewards.

\begin{figure}[t]
\centering
\includegraphics[width=0.9\columnwidth]{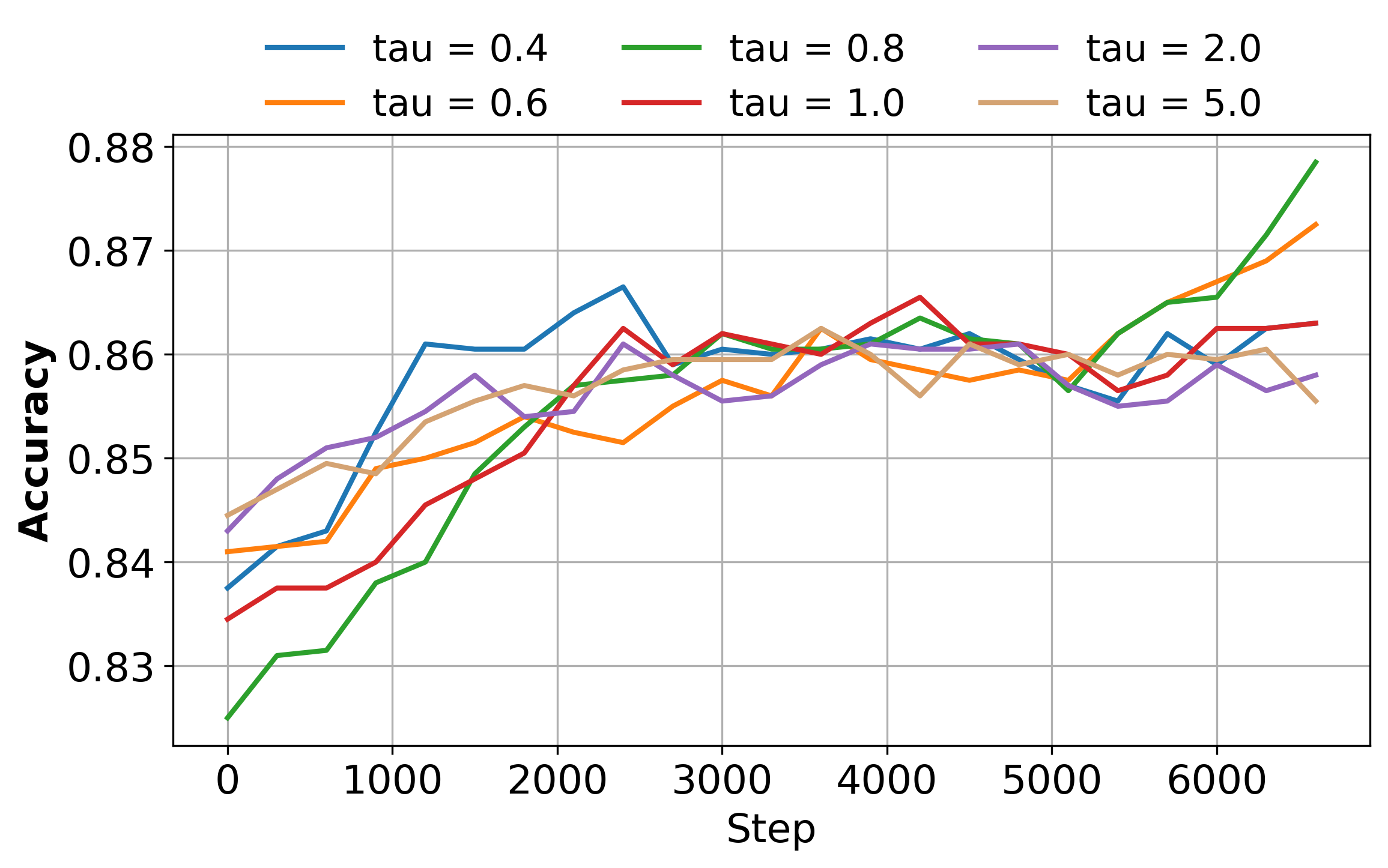}
\caption{Effect of parameter \(\tau\) on SoftRankPO performance. The method remains stable across a broad range of values, while finer tuning yields improved accuracy.}
\label{fig:tau_curve}
\end{figure}

\paragraph{Impact of Temperature Parameter \(\tau\) in SoftRankPO.}  
We conduct a controlled study varying the temperature coefficient \(\tau \in \{0.4, 0.6, 0.8, 1.0, 2.0, 5.0\}\), which adjusts the contrast in rank-derived advantage signals.
We observe stable convergence across all settings (Figure~\ref{fig:tau_curve}), confirming that SoftRankPO remains robust to a broad range of shaping intensities. Notably, setting \(\tau=0.8\) yields the best accuracy (approximately 88\%), improving upon the default \(\tau=0.5\) (86.35\%). 
Increasing \(\tau\) sharpens the distribution of transformed ranks, thereby amplifying differences in advantage estimates and enhancing gradient informativeness. Theoretically, since \(A_i^\star = \Phi^{-1}(p_i^\tau)\), a larger \(\tau\) spreads out the values of \(p_i^\tau\), especially near the extremes, increasing the gradient magnitude of the CDF inverse and leading to stronger learning signals for high-ranked actions. However, excessively low values of \(\tau\) compress the percentile spacing and may obscure meaningful preference distinctions. Thus, \(\tau\) 
influences both learning stability and policy expressiveness by regulating the granularity of rank sensitivity.


\begin{figure}[t]
\centering
\includegraphics[width=\columnwidth]{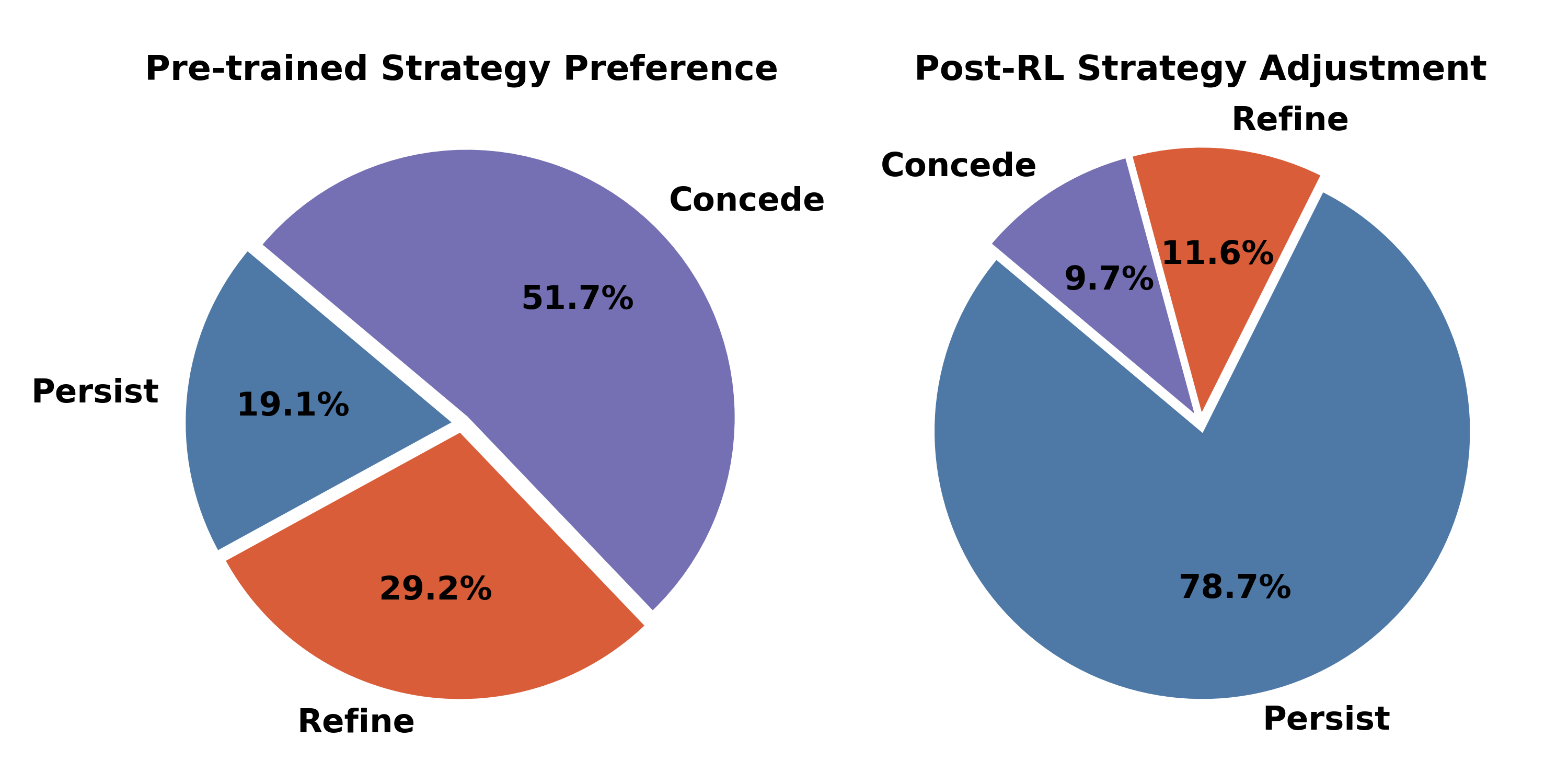}
\caption{Policy shift toward coordination: agents increasingly favor \texttt{PERSIST}, reflecting stronger confidence and selective intervention driven by reward shaping.}
\label{fig:action_policy}
\end{figure}

\paragraph{Policy Shift: From Collaboration to Coordination.}
\label{sec:policy_shift}
As shown in Figure~\ref{fig:action_policy}, SoftRankPO induces a significant behavioral shift: the \texttt{PERSIST} rate increases from 19.1\% to 78.8\%, while \texttt{REFINE} and \texttt{CONCEDE} jointly fall from 81\% to 21\%. This stark transition suggests that agents have learned to recognize and trust the high-quality outputs produced by the pretrained LLM. It reflects an emerging ability for self-assessment and confidence calibration. The resulting policy aligns with the principle of minimal intervention: instead of revising reflexively or adopting peer responses blindly, agents now act selectively, intervening only when they perceive a meaningful improvement. Such a pattern is directly encouraged by our reward shaping, which emphasizes marginal utility and penalizes redundant edits. Beyond accuracy, this leads to reduced token use (Table~\ref{tab:token_cost_small}). 

\section{Conclusion}
In this paper, we present MPDF, a meta-policy deliberation framework for multi-agent LLM collaboration, enabling agents to strategically adjust their behavior based on cognitive states. Coupled with the scale-resilient SoftRankPO algorithm, our approach achieves robust policy learning and fosters efficient coordination. 
Extensive results validate the effectiveness of our approach. 
We observe a policy shift during MPDF learning from collaboration to coordinatoin, marking a transition from exploratory language cooperation under SFT to coordinated decision-making under RL. 


\section{Acknowledgments}
This work was supported in part by a grant from the USC--Capital One Center for Responsible AI and Decision Making in Finance (CREDIF).

\bibliography{aaai2026}


\end{document}